# Using Discourse Signals for Robust Instructor Intervention Prediction

**Muthu Kumar Chandrasekaran[1], Carrie Demmans Epp[2,3], Min-Yen Kan[1,4], Diane Litman[2,5]**

[1] Department of Computer Science, School of Computing, National University of Singapore
[2] Learning Research and Development Center, University of Pittsburgh
[3] University Center for Teaching and Learning, University of Pittsburgh
[4] Interactive and Digital Media Institute, National University of Singapore, Singapore
[5] Department of Computer Science, University of Pittsburgh
{muthu.chandra, kanmy}@comp.nus.edu.sg     {cdemmans,dlitman}@pitt.edu

## Abstract

We tackle the prediction of instructor intervention in student posts from discussion forums in Massive Open Online Courses (MOOCs). Our key finding is that using automatically obtained discourse relations improves the prediction of when instructors intervene in student discussions, when compared with a state-of-the-art, feature-rich baseline. Our supervised classifier makes use of an automatic discourse parser which outputs Penn Discourse Treebank (PDTB) tags that represent in-post discourse features. We show PDTB relation-based features increase the robustness of the classifier and complement baseline features in recalling more diverse instructor intervention patterns. In comprehensive experiments over 14 MOOC offerings from several disciplines, the PDTB discourse features improve performance on average. The resultant models are less dependent on domain-specific vocabulary, allowing them to better generalize to new courses.

## Introduction

Massive Open Online Courses (MOOCs) aim to scale learning by creating virtual classrooms that eliminate the need for students to be co-located with instructional staff and each other. To facilitate interaction, MOOC platforms have discussion forums where students can interact with instructional staff – hereafter called *instructors* – and their classmates. Forums are typically the only mode of interaction between students and instructors. Forums often contain hundreds of posts from several thousand students, each post competing with others for instructor attention. Reading and responding to student queries in forums is an essential teaching activity that helps instructors gauge student understanding of course content. Intervention is argued to facilitate student learning where an instructor's presence and intervention in student discussions improves learning outcomes in MOOCs (Chen et al. 2016) and other online learning environments (Garrison, Anderson, and Archer 1999; Phirangee, Demmans Epp, and Hewitt 2016). However, instructors need to be selective when answering student posts due to their limited bandwidth. One selection strategy is to respond to posts that will maximally benefit the most students in a course. Along these lines, Chandrasekaran et al.

---

**Student 1 (original poster)**: Hie guys I m sorry **if**$_{Cont}$ my question is naive in anyway. **But**$_{Comp}$ I am confused ... Say suppose, **if**$_{Cont}$ we were to take the 5-6 Descending progression... **and so on**$_{Exp}$ I cant help **but**$_{Comp}$ see the ... **Now**$_{Temp}$ **if**$_{Cont}$ I need to apply the same progression to a minor scale, **then**$_{Cont}$ should I ... In the case of circle of fifths progression, **if**$_{Cont}$ I... **So**$_{Cont}$ we apply VII major instead?...

---

**Student 2 (1st reply)**: In a minor key the chords ... are **as follows**$_{Exp}$..., **but**$_{Cont}$ we ..., **because**$_{Cont}$ dominant chords should be... The wrinkle in minor keys is that to create the V chord, you have to raise the seventh degree of the scale (**so**$_{Cont}$ in a minor you sharp the g, **when**$_{Temp}$ it occurs in the V chord). I am not sure, **but**$_{Exp}$ I think ... I believe you can use that chord as an substitute for the V chord in a minor key, **just as**$_{Comp}$ you can in the major key, **but**$_{Exp}$ I'd depend on the staff to confirm that. ...Take this with a grain of salt, **as**$_{Cont}$ I am learning, too, **but**$_{Exp}$ I think it is correct.

---

**Instructor's reply**: Hi [Student1] [Student2] is heading towards the right direction. He is right about the circle of fifths ...

---

Figure 1: An example from our CLASSICAL-1 MOOC in our corpus where student confusion could benefit from instructor intervention. Here, discourse connectives are in bold and annotated with their Level-1 PDTB senses: (Temp)oral, (Cont)ingency, (Comp)arison, or (Exp)ansion.

(2015b) proposed an intervention taxonomy based on transactive discourse that details the situations in which certain types of interventions would maximally benefit students.

Consistent with this taxonomy, Chandrasekaran et al. showed that intervention strategies in MOOC forums differ widely. The factors behind different instructor intervention strategies include the instructors' pedagogical philosophy, a desire to encourage learner interaction, a desire to ensure students understand course content, and a need to correct misconceptions (Phirangee 2016). The intervention strategy chosen was found to impact student learning significantly (Mazzolini and Maddison 2003; 2007).

Earlier work also shows that content-based features, which include simple linguistic features derived from student vocabulary (e.g., word unigrams), signal common traits that are useful for predicting instructor interventions (Chandrasekaran et al. 2015a; Chaturvedi, Goldwasser, and Daumé III 2014). However, a key problem with surface-level vocabulary features is that they vary widely across courses, as courses from different subject areas use different domain-specific vocabulary. Predictive models trained on such word-based features do not generalize well when applied to new unseen courses in different disciplines. In contrast, function words such as conjunctions (e.g., "and", "because") occur frequently across corpora and can be leveraged to create more robust features, such as what was done in the related task of predicting transactivity in educational dialogues (Joshi and Rosé 2007).

Our work focuses on a specific subclass of function words — discourse connectives — as they serve to connect clauses and signal the communicative intent of the writer. The Penn Discourse Treebank (PDTB; Prasad et al. 2008) formalism identifies connectives that signal discourse relations and categorises them into senses. As can be seen in Figure 1, both student posts contain a number of *if...then*, *but* connectives that belong to the contingency and comparison senses of the PDTB. It is common to find such patterns in student posts expressing confusion as they hedge and hypothesise to check their understanding, which can call for instructor intervention. In contrast, the student post in Figure 3 is confident in tone, uses the imperative form, and is devoid of such connectives. These examples motivate us further to extract discourse-based features from student discussion forum posts. We hypothesize that student posts differ in their discourse structures and that some of these structures will attract instructor intervention. We additionally hypothesize discourse features will yield models for predicting instructor intervention that generalize well to unseen courses.

Following prior work, we cast the problem of predicting instructor intervention as a binary classification problem where intervened and non-intervened threads are treated as positive and negative instances, respectively. We test our hypotheses extrinsically using automatically extracted discourse features that follow the PDTB formalism, enriching a state-of-the-art baseline model for predicting instructor intervention. In contrast to prior work on single MOOC instances, our experiments are comprehensive, covering a corpus of 14 MOOC instances from various disciplines, offered by two different universities.

Our results show that PDTB features improve the state-of-the-art baseline performance by 3.4% (Table 4) when trained on a large out-of-domain dataset and by 0.4% (Table 3) when trained on a smaller in-domain dataset. Further, PDTB features on their own perform comparably to the state-of-the-art on select MOOC offerings. We show that unlike vocabulary based features, PDTB features are robust to domain differences across MOOCs.

## Related Work

Predicting instructor intervention became a viable problem to study with the availability of large amounts of educational discussion forum data from MOOCs. Chaturvedi et al. (2014) first specified the problem as predicting which MOOC discussion forum threads instructors would post to, where an instructor post is considered an intervention.

They modelled macro-level thread discourse structure (i.e., across posts), demonstrating that their model outperformed a representative classifier endowed with many lexical and other surface level features. However, later work failed to replicate their results across a much broader study of MOOC forums culled from several universities (Chandrasekaran et al. 2015a). This work cited the large variety of course content and instructor preference as likely causes to the non-portability of the initial study's results. Chandrasekaran et al. also showed that other simple features such as sub-forum type, thread length and surface level linguistic cues outperform the discourse model from the earlier work.

Our work uses Chandrasekaran et al. (2015a), hereafter denoted as EDM'15, as a starting point and as a state-of-the-art baseline for comparison. In contrast to both prior works, we model microscopic discourse structures – *i.e.*, sentence and clause-level discourse within student posts. We also eschew vocabulary-dependent approaches, such as those suggested by Ramesh et al. (2015) where intervention was based on emergent topics and subtopics from each course, since we seek models that generalize across a wide variety of courses.

**Discourse Parsing Applications.** As forum discussions feature dialogue and argumentation, we felt strongly that providing discourse analyses would improve prediction performance. Automatic discourse parsing discovers the relationship between clauses or sentences in contiguous text. Discourse parsing usually categorizes the inferred relation with a discourse type.

With the availability of large-scale discourse annotations on top of the Penn Treebank, the PDTB formalism for discourse annotation has become a *de facto* standard for automated discourse parsing and analyses. Importantly, the PDTB formalism splits the detection of discourse relations into ones signaled *explicitly* by a discourse connective (e.g., the connective "if" often signals a *Contingency* relation between its arguments, as in the first connective from Figure 1) from *implicitly* signaled ones that have no overt connective. As a result, automatic discourse relation identification relying solely on explicit connectives is rather precise but provides low overall coverage.

While the PDTB annotated corpus is built largely on newswire (e.g., *Wall Street Journal*), the PDTB tag set and derived parsers have found applicability in a variety of NLP tasks on different corpora: Li et al. (2014) showed the influence of PDTB explicit relations on machine translation quality. Mihăilă and Ananiadou (2014) found causal relations using PDTB in scientific Biomedical journals. Importantly, PDTB's applicability to the related form of user-generated text, especially expository texts, has also been studied: Faulkner et al. (2014) used PDTB discourse features to support argument classification in student essays from the International Corpus of Learner English[1]. Also similarly, in performing a selection problem close to ours, Wang et al. (2012) studied discourse parsing's utility for retweetability of tweets and found correlations between the discourse type and sentiment polarity. Swanson, Ecker, and

---

[1]`https://www.uclouvain.be/en-cecl-icle.html`

Walker (2015) also found these relations to be useful in argument extraction from general web forum text.

MOOC discussion forum text is user-generated, expository and conversational all at once. For this reason, we hypothesize that (explicitly marked) discourse parsing would improve the prediction of instructor intervention. Our hypothesis extends to forums in any online learning environment, such as the learning management systems that schools and universities host for their students.

## Data and Preprocessing

The corpus for our experiments consists of data from 14 Coursera MOOC offerings[2] that are spread across 7 courses from the authors' universities. The included MOOCs taught a variety of subjects spanning the humanities and sciences. All courses relied on videos to deliver core content. Different instructional approaches and learning activities (e.g., peer/self-assessments, prompted discussions, tests, or papers) were used and influenced discussion forum activities. This variety is apparent through the number of instructional staff (i.e., instructor or teaching assistants) who posted in the forums. The varied approaches are also apparent through their intervention ratios where CLINICAL-1[3] (Row 6; Table 1) had the highest intervention ratio (0.73) and DISASTER-3 (Row 10 of Table 1) had the lowest (0.02).

Coursera forums are divided into several sub-forums. Each of the sub-forums was manually categorized, using the definitions from (Chandrasekaran et al. 2015a), as belonging to one of the following types: errata, exam, lecture, homework, general, peer review, study group, or technical issues. Similar to prior work on intervention prediction, the general, study group, peer review and technical issues subforums and their threads were removed since they are noisy and do not focus on course content (e.g., social discussions and reports of technical issues). As the task is instructor intervention prediction, we also omit threads where the first post was made by an instructor. Table 1 shows the number of threads that are intervened in our model.

We truncate threads after the first instructor post (dropping subsequent student posts) because predicting the first instructor intervention is a viable problem and distinct from predicting subsequent, follow-up interventions that can be motivated by different reasons. Further, after an intervention, discussions gain visibility which can inflate feature counts in our prediction task. To extract features, we first tokenize thread text. We replaced instances of non-lexical references such as equations, URLs and timestamps, with the tokens: <EQU>, <URL>, <TIMEREF>, respectively. These tokens are a feature of the baseline prediction system (see "Baseline (EDM'15)" section). They also enable the discourse parser to skip unparsable text[4]. Stopwords and words of length less than 3 were removed before extracting the baseline features. Stopwords were not removed when extracting discourse features (cf "Discourse Feature Extraction" Section). Our work examines three predictive models: 1) the baseline (EDM'15), 2) a system with only PDTB discourse relations as features (PDTB), and 3) an augmented system where discourse relations are also used (EDM'15 + PDTB; E+P for short).

| Uni. | Course (-Iteration) | # of intervened | # of non–intervened | I. Ratio |
|---|---|---|---|---|
| NUS | CLASSIC-1 | 164 | 527 | 0.31 |
| | CLASSIC-2 | 17 | 155 | 0.11 |
| | REASON-1 | 58 | 231 | 0.25 |
| | REASON-2 | 40 | 265 | 0.15 |
| Pitt | ACCTALK | 98 | 254 | 0.39 |
| | CLINICAL-1 | 33 | 45 | 0.73 |
| | CLINICAL-2 | 32 | 82 | 0.39 |
| | DISASTER-1 | 81 | 2332 | 0.03 |
| | DISASTER-2 | 53 | 718 | 0.07 |
| | DISASTER-3 | 18 | 960 | 0.02 |
| | NUCLEAR-1 | 272 | 779 | 0.35 |
| | NUCLEAR-2 | 93 | 255 | 0.36 |
| | NUTRITION-1 | 98 | 2346 | 0.04 |
| | NUTRITION-2 | 73 | 1475 | 0.04 |
| | Total | 1,130 | 10,424 | |

Table 1: Thread counts over the four main sub-forums (errata, exam, lecture and homework) of each course iteration, with their intervention ratio (I. Ratio), defined as the ratio of # of intervened to non-intervened threads.

## Baseline (EDM'15)

The baseline system uses a maximum entropy classifier with the following set of features: unigrams, thread forum type, student affirmations to a previous post, thread properties (# of posts, comments, and posts+comments), average # of comments per post, # of sentences in the thread, # of URLs, and # of timestamped references to parts of a lecture video. The authors noted the imbalanced nature of their datasets, with non-intervened threads greatly outnumbering intervened ones. This motivated the use of class weights to counterbalance the # of non-intervened instances. Class weights, an important parameter of this model, were estimated as the ratio of negative to positive samples in the training instances.

## Discourse Feature Extraction

We experimented with a prediction system based solely on automatically-acquired discourse features from the PDTB-based discourse parser from (Lin, Ng, and Kan 2014). We employ this shallow discourse parser on the input to categorize identified discourse connectives according to the PDTB tag set, and subsequently extract them for our use. The parser first distinguishes discourse connectives (e.g., "and" can signal a discourse relation of 'Expansion', but can also act as a coordinating conjunction). It then classifies them into one of several senses as specified by PDTB. PDTB categorizes the connectives into implicit and explicit connectives, each of which is assigned a sense. Senses are organized hierarchically, where the top Level–1 senses discriminate among 4 relations: 'Contingency', 'Expansion', 'Comparison' and 'Temporal'. We only used explicit connectives and Level–1 senses as features. This is because (Lin, Ng, and Kan 2014) report a low $F_1$ of 39.6% for extracting implicit connectives while that of explicit connectives is much better (86.7%). Limiting to Level–1 senses also avoids sparsity issues. We found the distribution of the 4 Level–1 senses as tagged by the discourse parser in our MOOC corpus was similar to that of the original PDTB annotated corpus built from newswire (see Table 2), supporting our decision to use Level–1 senses.

| Corpus | Exp. | Cont. | Comp. | Temp. |
|---|---|---|---|---|
| **14 MOOC corpus** | 33% | 28% | 20% | 19% |
| **PDTB corpus** | 34% | 19% | 29% | 19% |

Table 2: Distribution of PDTB level-1 senses for explicit connectives (top row) as tagged by the discourse parser, which is similar to those reported for the PDTB corpus (Prasad et al. 2008), bottom row.

We used the Java version of the parser[5], which comes pre-trained with Sections 2–21 of the PDTB annotated corpus, using the Level–1 relation senses. We note that although the discourse parser's performance on MOOC forum text had not been previously evaluated, we decided to use the pre-trained parser given that such PDTB discourse parsers have been used to support a variety of downstream tasks using different corpora, without retraining (*cf* Related Work). Additionally, re-training is a resource intensive task, and we judged it to be a lower priority to evaluate it specifically for MOOC data.

Each forum thread is treated as a document, where each post in the thread is treated as a paragraph. Since the parser identifies discourse relations within paragraphs of text, only within-post discourse relations were identified; this is appropriate as the parser was trained on single-party narrative (newswire) rather than multiparty dialogue. We derive 25 features from the PDTB relation senses output by the parser. These constitute the discourse features identified as PDTB in Tables 3, 4 and 5:

- Total number of all relation senses (1 feature): The sum of the frequencies (number of occurrences) of all four Level–1 senses;

- Proportion of each sense (8 features): The absolute and relative frequency of each sense in a thread. Absolute frequency is normalized by thread length;

- Proportion of sense sequences of length 2 (16 features):

Normalized number of occurrences of each sense sequence of length 2 (e.g., 'Expansion'-'Contingency') in a thread divided by the total number of occurrences of all sense sequences of length 2.

We use the maximum entropy classifier with class weights as in the EDM '15 baseline for both PDTB and E+P systems. The implementations of the EDM '15 and the discourse based systems are publicly available[6].

## Evaluation

We evaluate the models under two evaluation schemes: an (i) in-domain scheme, and an (ii) out-of-domain scheme. We report the performance of the models in terms of precision (P), recall (R) and $F_1$ of the positive class.

The **in-domain setting** models were trained and evaluated separately on each MOOC using stratified five-fold cross validation. Stratification accounts for the highly imbalanced data and ensures that each fold had both positive and negative samples. We see that the combined model E+P outperforms the baseline EDM'15 (Table 3), on average.

Drilling down, we see that while EDM'15 does well on the first offering (those with the "-1" suffix) of many courses which have higher intervention ratios, E+P outperforms EDM'15 on subsequent offerings which typically have lower intervention ratios. We also observe the $F_1$ scores for CLASSIC-2 and DISASTER-3 are 0. Despite stratification, the # of intervened threads per fold was too low for both courses (~3 to 4 per fold) due to their low intervention ratios (see Table 1). As a result, both models are unable to predict any intervention for either course. In the **out-of-domain setting**, we use leave-one-out cross-course-validation (LOO-CCV) where models trained on 13 courses are tested on the 14[th] unseen course. This evaluation setting more closely approximates the real world where universities hosting MOOCs have data from previously offered MOOCs and would want to train predictive models that could be deployed in upcoming courses. This evaluation shows which models are more robust when adapting to unseen out-of-domain data. Table 4 shows the performance of the EDM'15 and E+P models on each of the 14 MOOCs from LOO–CCV.

E+P betters EDM'15 performance by 3.4% on average. The improved $F_1$ is largely due to a 5.7% improvement in recall. We argue that, for the problem of intervention prediction, improving recall is more important than precision since missing an intervention is costlier than intervening on a less important thread. Here the performance outlier is the CLINICAL-1 MOOC, where EDM'15 performs significantly better than E+P, which may be partially attributed to the course having the smallest test set.

Further, Tables 3 and 4 show that the E+P and EDM'15 models both benefit from access to more data in the out-of-domain setting. The EDM'15 model only improved by 2.7%, whereas the E+P model improved by 5.7%, showing the benefits of using domain-independent linguistic features to predict instructor intervention.

| Course | EDM'15 | | | EDM'15 + PDTB | | |
|---|---|---|---|---|---|---|
| | P | R | $F_1$ | P | R | $F_1$ |
| CLASSIC-1 | 25.0 | 33.1 | **28.5**** | 22.7 | 33.2 | 27.0 |
| CLASSIC-2 | 0.0 | 0.0 | 0.0 | 0.0 | 0.0 | 0.0 |
| REASON-1 | 32.2 | 48.2 | **38.6** | 25.6 | 41.8 | 31.8 |
| REASON-2 | 20.4 | 47.5 | 28.5 | 27.3 | 51.0 | **35.5** |
| ACCTALK | 59.3 | 44.7 | **51.0*** | 40.3 | 50.7 | 44.9 |
| CLINICAL-1 | 49.7 | 34.7 | **40.8** | 51.0 | 27.3 | 35.6 |
| CLINICAL-2 | 44.2 | 61.3 | 51.4 | 49.0 | 62.3 | **54.9** |
| DISASTER-1 | 14.7 | 6.7 | 9.2 | 16.0 | 9.4 | **11.8**** |
| DISASTER-2 | 6.7 | 5.0 | **5.7** | 6.7 | 3.3 | 4.4 |
| DISASTER-3 | 0.0 | 0.0 | 0.0 | 0.0 | 0.0 | 0.0 |
| NUCLEAR-1 | 15.5 | 16.8 | **16.1** | 14.3 | 12.9 | 13.6 |
| NUCLEAR-2 | 11.8 | 19.4 | 14.7 | 20.0 | 37.2 | **26.0**** |
| NUTRIT-1 | 85.5 | 57.8 | **69.0**** | 75.8 | 58.2 | 65.9 |
| NUTRIT-2 | 60.1 | 9.5 | 47.7 | 61.3 | 47.9 | **53.8*** |
| **Macro avg.** | 30.4 | 29.6 | 30.0 | 29.3 | **31.1** | 30.2 |
| **Weighted macro avg.** | 37.3 | 28.1 | 32.0 | 35.1 | **30.0** | 32.4 |

Table 3: Model performance of EDM'15, with and without PDTB, per MOOC, where each MOOC is evaluated individually (in-domain setting) using 5-fold stratified cross validation. Best performance is bolded; significance indicated where applicable ($^*p < 0.05$; $^{**}p < 0.01$).

| Course | EDM'15 | | | EDM'15 + PDTB | | |
|---|---|---|---|---|---|---|
| | P | R | $F_1$ | P | R | $F_1$ |
| CLASSIC-1 | 26.7 | 3.1 | 5.6 | 23.6 | 29.5 | **26.2**** |
| CLASSIC-2 | 18.5 | 31.3 | 23.3 | 18.2 | 37.5 | **24.5** |
| REASON-1 | 40.0 | 12.3 | 18.8 | 50.0 | 16.3 | **24.6** |
| REASON-2 | 52.9 | 29.0 | 37.5 | 35.6 | 51.6 | **42.1**** |
| ACCTALK | 41.0 | 26.7 | 32.3 | 50.0 | 26.7 | 34.8 |
| CLINICAL-1 | 81.8 | 30.0 | **43.9*** | 71.4 | 16.7 | 27.0 |
| CLINICAL-2 | 55.9 | 76.0 | 64.4 | 59.2 | 76.0 | **66.7** |
| DISASTER-1 | 25.6 | 14.5 | 18.5 | 21.8 | 25.0 | **23.3**** |
| DISASTER-2 | 20.0 | 4.8 | **7.8** | 18.8 | 4.8 | 7.7 |
| DISASTER-3 | 9.5 | 11.1 | 10.3 | 8.6 | 16.7 | **11.3**** |
| NUCLEAR-1 | 55.6 | 4.8 | 8.8 | 66.7 | 5.7 | **10.6** |
| NUCLEAR-2 | 33.3 | 15.6 | **21.2** | 31.8 | 15.6 | 20.9 |
| NUTRIT-1 | 77.3 | 62.4 | **69.1** | 72.0 | 63.4 | 67.4 |
| NUTRIT-2 | 46.5 | 52.4 | 49.3 | 54.2 | 50.8 | **52.5*** |
| **Macro avg.** | 41.8 | 26.7 | 32.6 | 41.6 | **31.2** | 35.6 |
| **Weighted macro avg.** | 42.7 | 29.3 | 34.7 | 41.9 | **35.0** | 38.1 |

Table 4: Prediction performance of EDM'15 and E+P systems in each of the 14 MOOCs where each one is evaluated (out-of-domain setting) using leave-one-out cross-course-validation (LOO–CCV). Best performance is bolded; significance indicated where applicable ($^*p < 0.05$; $^{**}p < 0.001$).

## Discussion

To understand the observed performance of the PDTB-based features, we probe further, answering two research questions that are natural extensions of the results.

**RQ1.** *Are the PDTB features useful supplemental evidence, especially when simple features do not perform well?*

In each of the 5 courses where E+P performs better than EDM'15, the course iterations have smaller intervention ratios (see Tables 1 and 3). For example, E+P betters EDM'15 on CLINICAL-2, REASON-2 and DISASTER-1 while EDM'15 has a better score on CLINICAL-1, REASON-1 and DISASTER-2. That is, PDTB features boost EDM'15 performance when there are fewer positive instances to learn from. This could be due to EDM'15's much larger feature space that requires more data to prevent sparsity. Note EDM'15 excluded stopwords, a subset of which are PDTB connectives, meaning that PDTB features contribute different information to the signal in the E+P model. Our analysis of the contributions of features showed 'Contingency' and 'Expansion' relations to contribute the most. This may be due to their higher prevalence relative to the other discourse relations in the corpus.

Consider the example in Figure 2. E+P classifies this thread correctly while the EDM'15 model fails. This short thread does not contain many content words. In contrast, the discourse connectives in these student posts activate 16 of the 25 PDTB features.

| |
|---|
| **Student 1 (Original poster)**: Hi !! I have a question about the 4th bar of the practice solution: the V chord has three roots. Is that normal **or**$_{Exp}$ just a mistake? Thank you. |
| **Student 2 (1st reply)**: [Student1's name], it is not a mistake. **While**$_{Comp}$ not as common **as**$_{Comp}$ merely doubling the root, **if necessary**$_{Cont}$, you can triple the root. **As**$_{Cont}$ you see in this case, the root is tripled to smooth out the voice leading. **Otherwise**$_{Comp}$, the tenor would be on the E, **and**$_{Exp}$ that would (probably) **result in**$_{Cont}$ parallel fifths moving to beat 4. |
| **Student 3**: Thanks [Student2's name] **but**$_{Comp}$, are you sure? I don't see it clear.... |
| **Instructor's reply**: Hi [Student3's name]. Please read these threads. It has been discussed before. Thanks. <URL> ... |

Figure 2: Both PDTB and E+P capture this intervention, while EDM'15 fails to capture this clarifying intervention.

While the use of explicit discourse connectives helped the classification task in many cases (e.g., Figure 2 and 4), the PDTB parser does not cover all of the observed discourse connectives or their expressed senses. Examples of connectives (in bold) that were not recognized include:

- ...not **as nice as** I thought it would be...
- *There's **only so much** melodic expressiveness... when using **nothing but** chord tones...*

**Student (Original poster)**: Try to search Epipen auto-injector. It is an epinephrine single-dose injection used to aid in severe allergic reaction or anaphylaxis. It's expiration date is around 1 year **after**$_{Temp}$ manufacturing date.

**Instructor's reply**: Be very careful concerning epipens. They are for severe reactions only ... They are prescription only for very good reasons as they affect the heart... It is probably bad for you ...

Figure 3: An instructor intervention to correct a student's misconception. Both the EDM'15 and E+P systems fail to predict this intervention.

| | EDM'15 | | | PDTB | | |
|---|---|---|---|---|---|---|
| **Course** | in | out | gain | in | out | gain |
| CLASSIC-1 | 28.5 | 5.6 | -22.9 | 22.9 | 32.5 | 9.5 |
| CLASSIC-2 | 0.0 | 23.3 | 23.3 | 14.5 | 21.9 | 7.4 |
| REASON-1 | 38.6 | 18.8 | -19.8 | 12.2 | 30.3 | 18.1 |
| REASON-2 | 28.5 | 37.5 | 9.0 | 21.9 | 33.7 | 11.8 |
| ACCTALK | 51.0 | 32.3 | -18.7 | 32.9 | 31.1 | -1.8 |
| CLINICAL-1 | 40.8 | 43.9 | 3.1 | 6.2 | 32.6 | 26.4 |
| CLINICAL-2 | 51.4 | 64.4 | 13.1 | 20.6 | 20.5 | -0.1 |
| DISASTER-1 | 9.2 | 18.5 | 9.3 | 8.9 | 4.0 | -4.8 |
| DISASTER-2 | 5.7 | 7.8 | 2.1 | 14.7 | 16.1 | 1.4 |
| DISASTER-3 | 0.0 | 10.3 | 10.3 | 2.0 | 7.1 | 5.2 |
| NUCLEAR-1 | 16.1 | 8.8 | -7.3 | 10.8 | 23.9 | 13.0 |
| NUCLEAR-2 | 14.7 | 21.2 | 6.5 | 21.7 | 20.3 | -1.4 |
| NUTRIT-1 | 69.0 | 69.1 | 0.1 | 6.7 | 9.9 | 3.18 |
| NUTRIT-2 | 47.7 | 49.3 | 1.6 | 9.7 | 14.8 | 5.11 |
| **Macro avg.** | 30.0 | 32.6 | **2.6** | 16.4 | 25.3 | **8.9** |
| **Weighted macro avg.** | 32.0 | 34.7 | **2.7** | 11.8 | 23.4 | **11.6** |

Table 5: Prediction performance of EDM'15 and PDTB systems for each of the 14 MOOCs in the (in)-domain and (out)-of-domain evaluations.

The presence of these connectives in our data is consistent with recent calls (Forbes-Riley, Zhang, and Litman 2016) to modify the PDTB relation inventory by adding a broader set of tags, such as those suggested by Tonelli et al. (2010). Increasing the PDTB's coverage of explicit connectives would likely improve results. The added use of implicit connectives may also improve prediction performance, should implicit connective detection and classification be improved. Figure 4 shows an example where implicit connectives may strengthen signals from discourse features. We also note that there are cases, such as in Figure 3, which lack discourse and lexical signals. The intervention here is instead triggered by domain knowledge. These excerpts exemplify the difficulty of our prediction task.

**RQ2.** *Are PDTB features more robust than vocabulary-based features?*

Consider the performance differences of the EDM'15 and PDTB models between the in-domain and out-of-domain

**Student 1 (Original Poster)**: Well, hurricanes are #1 in the summer time. **/So/**$_{Cont}$ I always have a hurricane kit handy -3 days worth of supplies. **/Also/**$_{Exp}$,Floods, especially coastal and flash flooding are prevalent. Tornadoes, severe storms with damaging winds and hail. **/But/**$_{Comp}$ We don't have to worry about snow **or**$_{Exp}$ fires too much. Earthquakes, either, **although**$_{Comp}$ we do have a fault line nearby. **/so far/** We've had a couple of very minute tremors in my lifetime, **but**$_{Exp}$ nothing that is really noticed.

**Student 2**: Tornadoes and hail/wind storms are the most prevalent disaster to my area, **although**$_{Comp}$ we have been hit with just about everything, including a hurricane (Hurricane Ike took our roof!)! Flooding is an issue in this area, **as**$_{Exp}$ is extreme winter weather on occasion; in recent years, significant snowfall (2 feet) in a short period of time basically paralyzes communities such **as**$_{Comp}$ ours that do not contend with such very often. We have **also**$_{Exp}$ had significant ice storms that have caused incredible damage. We are well aware of the fact that we could get a significant earthquake, **though**$_{Comp}$ fortunately we have had only minor issues in that regard...

**Instructor's reply**: I was involved in some of the response after Katrina... I met some amazing folks and saw some real devastation. One thing I never got used to ... Still I was glad to be there and the people - amazing ...

Figure 4: An instructor intervention to build common ground with students. The PDTB model predicts this intervention, while EDM'15 and E+P fail. Also shown in the first post are *implicit* connectives, within /../, that are not captured.

evaluation settings (see Table 5). Using PDTB features results in an average improvement of 11.6% when evaluating models out-of-domain, whereas EDM'15 only improves by 2.7%. EDM'15 performance drops greatly on CLASSIC-1, REASON-1, and ACCTALK due to the out-of-domain data; in contrast, PDTB gains on 10 of the 14 courses. In the example in Figure 4, EDM'15 and E+P fail while PDTB predicts correctly. Frequent words from this course (e.g., "tornado", "earthquake") are rare across MOOCs; this weakens the EDM'15 model.

These findings suggest that PDTB may result in further gains were data added from more courses, while that of the vocabulary-based EDM'15 model may worsen or not scale. Similarly, Chandrasekaran et al. (2015a) did not see improvements in the EDM'15 model when they went from a 13 course training set to 60 courses. This demonstrated lack of robustness in the EDM'15 model is not surprising given its abundant use of domain-specific vocabulary. There were considerably more out-of-domain unigram features (76,382) than in-domain unigram features (15,161, on average). This steep increase in feature space and resulting sparsity hampers the EDM'15 model's ability to benefit from a scaled corpus. In contrast, both in-domain and out-of-domain versions of the PDTB model have the same number of features.

**Potential Improvements.** The above results indicate that performance improvements due to PDTB features scale better than vocabulary-based features. To harness similar improvements from large scale data, the performance and ro-bustness of the discourse parser needs to be improved. Current limitations of discourse parsers (e.g., their inability to process equations and symbols, lack of near real-time output) inhibit scaling the predictive power of PDTB and E+P models. Discourse parser improvements would therefore enable the development and use of better predictive models.

## Conclusion

In this study, we better the state-of-the-art for intervention prediction by augmenting it with PDTB relation based features. Further, on select MOOC offerings PDTB relations alone performed comparably to the state-of-the-art. Unlike vocabulary based models, PDTB based features were shown to be robust to domain differences across MOOCs. This domain independence supports the improved prediction of instructor interventions. The current $F_1$ scores are still markedly low. Better modelling of the instructor may help boost performance. We plan to tackle this in two ways. First, we will model instructor intervention based on the threads they have seen because they cannot intervene in threads that they have not seen. Second, we will model intervention based on the role that different types of instructional staff play. We expect teaching assistants, course alumni (also known as "community TAs" and "mentors"), and faculty to have different motivations for intervening. They may also dedicate different amounts of time to course forums. As a result, modelling these factors or individual instructor preferences could improve prediction performance.

## Acknowledgments

This research is funded in part by NUS Learning Innovation Fund – Technology grant #C-252-000-123-001, and by the Singapore National Research Foundation under its International Research Centre @ Singapore Funding Initiative and administered by the IDM Programme Office, and by an NUS Shaw Visiting Professor Award. The research is also funded in part by the Learning Research and Development Center at the University of Pittsburgh and by a Google Faculty Research Award. We would like to thank the University of Pittsburgh's Center for Teaching and Learning, for sharing their MOOC data.